\documentclass[11pt]{article}

\usepackage[preprint]{acl}

\usepackage{times}
\usepackage{latexsym}
\usepackage{tcolorbox}
\tcbuselibrary{breakable}
\usepackage{dblfloatfix}

\usepackage[T1]{fontenc}
\usepackage[utf8]{inputenc}

\usepackage{microtype}

\usepackage{inconsolata}

\usepackage{graphicx}

\usepackage{amsmath}
\usepackage{amssymb}
\usepackage{booktabs}
\usepackage{enumitem}
\usepackage{cleveref}
\usepackage{siunitx}

\title{Learning Perspectivist Social Meaning\\ via Demographic-Conditioned Fusion Embeddings}

\author{
  Amanda Cercas Curry \\
  Independent Researcher \\
  \texttt{amanda.cercas@gmail.com} \\\And
  Lucio La Cava \\
  University of Calabria \\
  \texttt{lucio.lacava@dimes.unical.it} \\ \AND
  Luca Maria Aiello \\
  IT University of Copenhagen \\
  \texttt{luai@itu.dk} \\ \And
  Gianmarco De Francisci Morales \\
  CENTAI \\
  gdfm@acm.org \\
}

\begin{document}
\maketitle
\begin{abstract}
Social meaning in language is inherently perspectival, varying across annotator backgrounds, demographics, and ideological positions.
However, most NLP systems collapse this variation into a single ground-truth label, ignoring the diversity of interpretations.
In this work, we model social dimensions along a \emph{perspectivist spectrum}, capturing how interpretations vary across demographic groups on a dataset consisting of 28k human annotations.
We benchmark multiple modeling paradigms, including zero-shot, few-shot, and fine-tuned approaches, and propose \textbf{fusion embeddings} that integrate textual and demographic representations.
Our fusion models yield consistent and statistically significant improvements over text-only baselines across all fusion strategies (+$5.9$--$6.5$\% relative macro PR-AUC), with shuffle ablations confirming that demographic profiles carry genuine predictive signal rather than spurious correlations.
\end{abstract}

\section{Introduction}
\label{sec:intro}

Social meaning is not fixed in text: it is constructed in the act of reading.
The same utterance can convey knowledge-sharing to one reader, status-conferral to another, and support to a third, depending on who interprets it and through which cultural, demographic, and ideological lens.
This perspectival nature of pragmatic meaning has long been acknowledged in sociolinguistics~\cite{plank-2022-problem,hovy2021importance}.
Yet, most NLP systems treat social labels as objective properties of text by collapsing the diversity of human interpretation into a single ground truth.

Building on decades of social science research, \citet{choi-etal-2020-ten} operationalized ten fundamental dimensions of social interaction---knowledge, power, status, trust, support, romance, similarity, identity, fun, and conflict---and showed that they can be reliably detected from conversational text.
Their framework has since been applied to study agreement, coordination, and community well-being in online settings~\cite{monti-etal-2022-language,lucchini-etal-2022-reddit,aiello-etal-2021-epidemic,balsamo-etal-2023-pursuit}.
However, the original dataset was collected with a homogeneous annotator pool, released only in aggregated form, and provided no information on annotator demographics, making it impossible to study how interpretation varies across people.

\textsc{P1SCO}~\citep{cercas-curry2026p1sco} addresses this gap by introducing a large-scale, disaggregated dataset of social dimension annotations across three social media platforms (Reddit, YouTube, and Instagram), contributed by 543 demographically diverse participants from the US and UK.
\textsc{P1SCO} demonstrates that gender, age, nationality, and political orientation, as well as Big Five personality traits, correlate with label assignment across all ten dimensions.
Crucially, homogeneous demographic groups exhibit higher within-group agreement than the overall population, thus suggesting that shared social experiences produce convergent interpretive frameworks.
These findings establish a key premise for the present work: annotator disagreement in this task is not noise, rather it is signal.

In this paper, we ask: \textit{can models learn to predict how different demographic groups perceive social dimensions?}
We frame this as a \textit{perspectivist prediction} task, in which the model receives both a candidate text and an annotator demographic profile, and must estimate the probability that an annotator with that profile would assign each social dimension.
Using \textsc{P1SCO} as our evaluation platform, we benchmark a range of modeling paradigms (zero-shot, few-shot, and fine-tuned) and propose \textbf{demographic-conditioned fusion embeddings} that integrate textual and demographic representations at multiple levels of depth.

Empirically, we find that demographic conditioning consistently improves over text-only baselines, with our best fusion model achieving a ${\sim}6.5\%$ relative gain in macro PR-AUC.
A shuffle ablation confirms that these gains reflect genuinely informative demographic signals rather than spurious correlations.
Per-label analysis reveals the largest improvements for semantically ambiguous dimensions, Power ($+51.9\%$ relative) and Trust ($+30.1\%$ relative), where demographic perspective most strongly mediates interpretation.

\paragraph{Contributions}
\begin{itemize}[leftmargin=*, nosep]
    \item We establish the first comprehensive baselines for perspectivist social dimension prediction on \textsc{P1SCO}, benchmarking zero-shot, few-shot, and fine-tuned paradigms.
    \item We propose fusion embeddings combining textual and demographic representations at three integration depths, achieving up to $+6.5\%$ relative macro PR-AUC over text-only models, with all fusion strategies yielding statistically significant gains.
    \item Through shuffle ablations and per-label analysis, we show that demographic signals carry genuine predictive information and that the full gender--age--nationality triplet outperforms any demographic subset alone.
\end{itemize}

\section{Task Formulation}
\label{sec:task}

Let $\mathcal{L}$ denote the ten-label set.
For a candidate $c$ with text $x_c$, let $R_c$ denote the set of annotators who labelled it, and let $v^{(r)}_{c,\ell} \in \{0,1\}$ be the binary label for $c$ and dimension $\ell \in \mathcal{L}$ assigned by $r \in R_c$, indicating whether a social dimension is present.

For any non-empty annotator subset $S_c \subseteq R_c$, we define a \textit{soft} classification target as the fraction of annotators in $S_c$ that recognized dimension $\ell$ in candidate text $c$ as:
\begin{equation}
  s_{c,\ell}(S_c) = \frac{1}{|S_c|}\sum_{r\in S_c} v^{(r)}_{c,\ell},
\end{equation}
and the corresponding \textit{hard} classification target as: 
\begin{equation}
  y_{c,\ell}(S_c) = \mathbb{I}[s_{c,\ell}(S_c) > 0.5].
\end{equation}

Specifically, $s_{c,\ell}(S_c)$ indicates the degree of agreement inside the annotator set $S_c$, whereas $y_{c,\ell}(S_c)$ indicates the corresponding binary majority-version (with exact ties defaulting to 0).

Note that, hereinafter, the data splitting is always performed at the candidate level using \textit{iterative multilabel stratification}, so as to avoid leaking candidate content between splits.

\paragraph{Majority Prediction}
This task corresponds a special case where $S_c=R_c$, i.e., all annotators are taken into account.
For this task, we derive the $s_{c,\ell}^{maj}=s_{c,\ell}(R_c)$ and $y_{c,\ell}^{maj}=y_{c,\ell}(R_c)$ majority labels by only considering candidates having at least three annotations.
The input is $x_c$, and the output yields one probability value per social dimension.
In the hard-label setting, we use $y_{c,\ell}^{maj}$ for evaluation, while in the soft-label setting we compare the model predictions with $s_{c,\ell}^{maj}$.

The final split for this task contains \num{3879}/\num{550}/\num{1104} train/val/test candidates.

\paragraph{Perspectivist Prediction}
This task setting defines the annotator subset $S_c$ based on demographic profiles.
Let us denote with $\mathbf{m}_r=(g_r,a_r,n_r)$ the demographic profile for annotator $r$, consisting of $r$'s gender, age group, and nationality, and let $G=(g,a,n)$ define a demographic group as a tuple.
For a candidate $c$, the group-specific annotator subset is:
\begin{equation}
R_c(G)=\{r\in R_c : \mathbf{m}_r = G\}.
\end{equation}

If $R_c(G)$ is non-empty, we can use the definitions above to define the corresponding group-specific soft and hard labels as $s_{c,\ell}^{(G)}=s_{c,\ell}(R_c(G))$ and $y_{c,\ell}^{(G)}=y_{c,\ell}(R_c(G))$.

We emphasize that, instead of collapsing individual annotations to group-level ones, we retain one record per observed annotation, exploiting intra-group variability to better calibrate predictions.
Specifically, task input corresponds to a pair $(x_c, \mathbf{m}_r)$ and the expected output estimates $p(v^{(r)}_{c,\ell}=1 \mid x_c,\mathbf{m}_r)$ for each label.

Note that for a given group $G$, this probability corresponds to an estimate of $s_{c,\ell}^{(G)}$, whereas the corresponding hard-label can be obtained by thresholding these probabilities.

The final split for this task contains \num{19022}/\num{2691}/\num{5480} train/val/test candidates.

\section{Architecture}
\label{sec:architecture}

\subsection{Text Encoder}
The main architecture we train to address the social dimension classification tasks consists of a fine-tuned RoBERTa-large~\cite{roberta}, implemented via the \textit{HuggingFace} library.
Let us denote with $E$ the encoder, with a set of parameters $\theta$.
For any candidate input text $x_c$, we collect the final-layer, first-token representation $\mathbf{h}_c=E_{\theta}(x_c) \in \mathbb{R}^H$, i.e., the $[CLS]$ token, with $H=1024$ for RoBERTa-large.

A small classification head maps this text representation to one logit per social dimension:
\begin{equation}
  \mathbf{z}^{\mathrm{text}}_c
  = C_{\mathrm{text}}(\mathbf{h}_c)
  \in\mathbb{R}^{|\mathcal{L}|}
\end{equation}
where $C_{\mathrm{text}}$ denotes the standard RoBERTa classification-head design.
Majority-vote social-dimension prediction leverages these logits directly, whereas perspectivist prediction uses them as a text-only baseline.

\subsection{Demographic Encoder}
\label{sec:demo-encoder}
To encode a given socio-demographic perspective, we consider the annotator's gender ($g$), age group ($a$), and nationality ($n$) and encode them separately into 64-dimensional vectors.
The encoding uses RoBERTa to transform the textual representation of the specific sociodemographic group into its embedding.
Then, we project the concatenated representation as follows:
\begin{align}
  \mathbf{d}_r &= D_{\mathrm{demo}}([\mathbf{e}^{g}(g_r);\mathbf{e}^{a}(a_r);\mathbf{e}^{n}(n_r)])
  \in\mathbb{R}^{128},
\end{align}
where $D_{\mathrm{demo}}$ is a two-layer MLP with GELU activation.

\subsection{Text-Perspective Fusion Modalities}
To integrate socio-demographics within our latent representations and obtain the perspectivist social dimension classification, we devise different fusion strategies acting at three depths, plus a baseline.

\paragraph{Text-only baseline}
This approach simply leverages the textual encoding $\mathbf{z}^{text}$, ignoring $\mathbf{d}_r$.
Note that, since we discard the sociodemographic conditioning, any improvement over this baseline suggests that sociodemographic integration contributes to prediction, beyond textual content.

\paragraph{Additive fusion}
Predictions under this fusion modality are obtained as:
\begin{equation}
  \mathbf{z}^{\mathrm{add}}_{c,r}
  = \mathbf{z}^{\mathrm{text}}_c + W_d\mathbf{d}_r + \mathbf{b}_d,
\end{equation}
where $W_d$ and $\mathbf{b}_d$ are zero-initialized, so the model starts demographic-blind and learns group-conditioned residuals only where the training data support them.
This represents the most conservative fusion strategy, as text remains the primary evidence, while demographics can only adjust the label logits.

\paragraph{Early fusion}
This modality concatenates the pooled textual representation and socio-demographic encodings before classification:
\begin{equation}
  \mathbf{z}^{\mathrm{early}}_{c,r}
  = C_{early}([\mathbf{h}_c;\mathbf{d}_r]),
\end{equation}
where $C_{early}$ follows the classification head of $C_{text}$ but receives as input both the text vector and the demographic vector.
Note that this fusion strategy can learn deeper, non-additive, text-demographic interactions.

\paragraph{Concat-then-encode fusion}
This represents the deepest integration strategy, as it prepends the demographic tuple to the text before tokenization and encoding, such as:
\[
\texttt{[Female, 25-34, UK] }\langle\text{text}\rangle .
\]

After concatenation, text is processed through the text-only encoder; thus this modality does not use the encoder described in \Cref{sec:demo-encoder}.
Here, the deepest integration stems from the possibility for every Transformer layer to attend the ``conditioning'' prefix while encoding the text.

\subsection{Training}
Given the prediction settings in \Cref{sec:task} and the architectures above, training differs only in which target vector is paired with each model output.
All models produce one logit and one probability per social dimension.
We write $\mathbf{z}_c\in\mathbb{R}^{|\mathcal{L}|}$ for candidate-level logits and  $\mathbf{p}_c=\sigma(\mathbf{z}_c)\in[0,1]^{|\mathcal{L}|}$ for the corresponding probabilities. 
Thus, $p_{c\ell}$ is the model probability that
social dimension $\ell$ applies to candidate $c$.
For models that work on disaggregated single-annotation labels, we analogously write $\mathbf{z}_{c}^{(r)}$ and $\mathbf{p}_{c}^{(r)}=\sigma(\mathbf{z}_{c}^{(r)})$, where $p_{c,\ell}^{(r)}$ estimates the probability that an annotator with profile $\mathbf{m}_r$ assigns label $\ell$ to candidate $c$.

\paragraph{Majority Social Dimension Prediction}
For the candidate-level model, the input is the text $x_c$.
We train three variants with the same probability output $\mathbf{p}_c$: (i) a hard-label one targeting $y_c^{maj}$, (ii) a soft-level variant targeting $s_c^{maj}$, and (iii) a variation of the latter focusing on MSE/Brier scores to directly optimize probability errors.

\paragraph{Perspectivist Social Dimension Prediction}
For the annotation-level prediction, the input is $(x_c,\mathbf{m}_r)$ and the model outputs  $\mathbf{p}_{c}^{(r)} \in [0,1]^{|\mathcal{L}|}$, where each component estimates $P(v_{c,\ell}^{(r)}=1\mid x_c,\mathbf{m}_r)$.

As mentioned in \Cref{sec:task}, during training, we use the observed annotation vector $\mathbf{v}_{c}^{(r)}$ derived from individual annotations $v_{c,\ell}^{(r)}$ as the target.
Conversely, at inference, fixing $\mathbf{m}_r=G$ gives a group-conditioned prediction $\hat{\mathbf{s}}^{(G)}_c=\mathbf{p}^{(G)}_c$, which serves as an estimate of the expected soft perspective of group $G$ on text $c$.
Note that the corresponding hard prediction is $\hat{y}^{(G)}_{c,\ell}=\mathbb{I}[\hat{s}^{(G)}_{c,\ell}>0.5]$.
Finally, our text-only baseline uses the same annotation-level target but omits $\mathbf{m}_r$, so it cannot produce group-specific predictions.

\paragraph{Objective Functions}
Let $\mathrm{BCE}_w$ denote multilabel binary cross-entropy with positive-label weighing, which accounts for rare labels (we similarly account for underrepresented demographics via sampling).
The aforementioned training objectives can be summarized as follows:
\begin{align}
  \mathcal{J}_{\mathrm{hard}}
    &= \mathrm{BCE}_w(\mathbf{y}^{\mathrm{maj}}_c,\mathbf{p}_c),\\
  \mathcal{J}_{\mathrm{soft}}
    &= \mathrm{BCE}_w(\mathbf{s}^{\mathrm{maj}}_c,\mathbf{p}_c),\\
  \mathcal{J}_{\mathrm{mse}}
    &= \frac{1}{|\mathcal{L}|}\sum_{\ell\in\mathcal{L}}
       (p_{c,\ell}-s^{\mathrm{maj}}_{c,\ell})^2,\\
  \mathcal{J}_{\mathrm{persp}}
    &= \mathrm{BCE}_w(\mathbf{v}_{c}^{(r)},\mathbf{p}_{c}^{(r)}).
\end{align}
Note that candidate-level models differ by whether they learn from hard consensus labels or soft annotation fractions, while annotation-level models learn from individual annotator votes, and their probabilities are interpreted as estimates of the expected group-level soft perspective when a demographic profile is provided.

\section{Experimental Setup}
\label{sec:experiments}

We evaluate on \textsc{P1SCO}, a dataset with social labels and annotator demographic information~\citep{cercas-curry2026p1sco}.

\subsection{Evaluation}
\paragraph{Majority Setting}
Each test sample is a candidate $c$.
The model outputs the probability vector $\mathbf{p}_c$ over the social dimensions, and we evaluate these probabilities against two candidate-level targets: the hard majority vector $\mathbf{y}^{\mathrm{maj}}_c$ and the soft annotation-fraction vector $\mathbf{s}^{\mathrm{maj}}_c$.

\paragraph{Perspectivist Setting}
Each test sample is a candidate-annotator pair $(c,r)$.
The model receives the text $x_c$ and the annotator profile $\mathbf{m}_r$, and outputs $\mathbf{p}_{c}^{(r)}$. 
This vector estimates how an annotator with profile $\mathbf{m}_r$ would label $x_c$, and evaluation compares it with the observed individual label vector $\mathbf{v}_{c}^{(r)}$.
A group-level perspective can be obtained at inference-time by fixing a profile $G$ and reading $\mathbf{p}^{(G)}_c$ as the expected soft perspective of that group.

\paragraph{Metrics}
We use macro PR-AUC as the primary metric under both settings, computed against the hard targets, i.e., $\mathbf{y}^{\mathrm{maj}}_c$ for the majority setting, and $\mathbf{v}_{c}^{(r)}$ for the perspectivist one.
This chioce ensures giving equal weight to each social dimension, is threshold-free, and is particularly suitable for sparse, multilabel data.
For the majority setting, we also report the Brier error against soft targets, whereby lower values indicate model probabilities better match the observed annotator agreement.

To compare the various perspectivist encodings, we also report PR-AUC confidence intervals over \num{1000}-sample bootstraps over candidates.

\section{Results}
\label{sec:results}

\begin{table}[t]
\centering
\small
\setlength{\tabcolsep}{3pt}

\begin{tabular}{llcc}
\toprule
Objective & Target & PR-AUC $\uparrow$ & Brier $\downarrow$ \\
\midrule
Hard BCE & Majority & 0.405 $\pm$ 0.011 & 0.0455 \\
Soft BCE & Fractions & \textbf{0.438 $\pm$ 0.009} & 0.0402 \\
MSE/Brier & Fractions & 0.438 $\pm$ 0.008 & \textbf{0.0295} \\
\bottomrule
\end{tabular}
\caption{Majority-task results. PR-AUC is macro PR-AUC against majority labels (mean $\pm$ std over three seeds). ``Fractions'' denotes annotation-fraction supervision, and Brier evaluates probability fit to annotation fractions; lower is better.}
\label{tab:task_a}
\end{table}

\begin{table}[t]
\centering
\small
\setlength{\tabcolsep}{3pt}
\resizebox{\columnwidth}{!}{%
\begin{tabular}{lccc}
\toprule
Model & PR-AUC & $\Delta$ vs text (pp) & 95\% CI \\
\midrule
Text-only        & 0.389 $\pm$ 0.001          & ---              & ---             \\
Additive         & \textbf{0.414 $\pm$ 0.002} & +2.53 (+6.5\%)$^\dagger$ & [+1.45, +3.70]  \\
Early Fusion     & 0.413 $\pm$ 0.001          & +2.34 (+6.0\%)$^\dagger$ & [+1.20, +3.56]  \\
Concat-Encode & 0.412 $\pm$ 0.003          & +2.30 (+5.9\%)$^\dagger$ & [+1.15, +3.49]  \\
\bottomrule
\end{tabular}%
}
\caption{Perspectivist-task macro PR-AUC (mean $\pm$ std over three seeds). $\Delta$ and CIs are absolute PR-AUC deltas in percentage points; relative gains are shown in parentheses. $\dagger$ = 95\% CI excludes zero.}
\label{tab:task_b}
\end{table}

\begin{table}[t]
\centering
\small

\resizebox{\columnwidth}{!}{%
\begin{tabular}{lccl}
\toprule
Label & Text-only & Additive & $\Delta$ vs text \\
\midrule
Power & 0.132 & 0.200 & +0.068 (+51.9\%) \\
Trust & 0.157 & 0.205 & +0.047 (+30.1\%) \\
Fun & 0.404 & 0.440 & +0.036 (+8.9\%) \\
Identity & 0.269 & 0.296 & +0.026 (+9.8\%) \\
Status & 0.412 & 0.436 & +0.024 (+5.9\%) \\
Similarity & 0.434 & 0.456 & +0.022 (+5.0\%) \\
Conflict & 0.561 & 0.583 & +0.022 (+3.9\%) \\
Support & 0.607 & 0.620 & +0.013 (+2.1\%) \\
Knowledge & 0.690 & 0.701 & +0.010 (+1.5\%) \\
\textit{Romance} & 0.226 & 0.209 & \textit{-0.017 (-7.3\%)} \\
\bottomrule
\end{tabular}%
}
\caption{Per-label PR-AUC (mean), sorted by absolute additive-minus-text gain. Relative gains are shown in parentheses. \textit{Italic} row = negative gain.}
\label{tab:per_label}
\end{table}

\begin{table}[t]
\centering
\small

\begin{tabular}{lccc}
\toprule
Model & Normal & Shuffled & $\Delta$ shuffled \\
\midrule
Additive & 0.414 & 0.384 & $-$0.030 ($-$7.2\%) \\
Early Fusion & 0.413 & 0.382 & $-$0.031 ($-$7.3\%) \\
\bottomrule
\end{tabular}
\caption{Shuffle ablation averaged over three seeds. ``Normal'' and ``Shuffled'' report macro PR-AUC with the full G+A+N triplet. $\Delta$ is the absolute shuffled-minus-normal difference, with relative drop in parentheses.}
\label{tab:shuffle}
\end{table}

\begin{table}[t]
\centering
\small

\begin{tabular}{lcc}
\toprule
Demographic subset & PR-AUC & $\Delta$ vs full \\
\midrule
\textbf{All (G+A+N)} & \textbf{0.414} & --- \\
Age + Nat. & 0.397 & $-$0.017 \\
Gender + Age & 0.396 & $-$0.018 \\
Age only & 0.396 & $-$0.018 \\
Gender + Nat. & 0.395 & $-$0.020 \\
Gender only & 0.394 & $-$0.021 \\
Nationality only & 0.394 & $-$0.021 \\
\bottomrule
\end{tabular}
\caption{Additive model macro PR-AUC by demographic subset (mean over 3 seeds). G=gender, A=age, N=nationality.}
\label{tab:subset}
\vspace{-.6\baselineskip}
\end{table}

\begin{table}[t]
\centering
\small

\begin{tabular}{lccc}
\toprule
Label & Best subset & Best PR-AUC & $\Delta$ vs all \\
\midrule
Knowledge & All (G+A+N) & 0.701 & --- \\
Power      & All (G+A+N) & 0.200 & --- \\
Status     & All (G+A+N) & 0.436 & --- \\
Trust      & All (G+A+N) & 0.205 & --- \\
Support    & All (G+A+N) & 0.620 & --- \\
Similarity & All (G+A+N) & 0.456 & --- \\
Identity   & All (G+A+N) & 0.296 & --- \\
Fun        & All (G+A+N) & 0.440 & --- \\
Conflict   & All (G+A+N) & 0.583 & --- \\
\textit{Romance} & \textit{Gender only} & \textit{0.225} & \textit{+0.015} \\
\bottomrule
\end{tabular}
\caption{Most informative demographic subset per label (additive, mean over 3 seeds). 9/10 labels: full triplet is best. \textit{Italic}: romance uniquely benefits from gender alone (+1.5pp over full triplet).}
\label{tab:best_subset}
\vspace{-.6\baselineskip}
\end{table}

\Cref{tab:task_a} reports Majority Prediction results across the three supervision objectives.
Replacing hard majority-vote labels with soft annotation fractions (Soft BCE) yields a $+3.3$ pp improvement in macro PR-AUC ($0.438 \pm 0.009$ vs.\ $0.405 \pm 0.011$). %
This result confirms that annotator disagreement carries predictive signal beyond what majority-vote labels encode.
The MSE/Brier objective matches Soft BCE on PR-AUC ($0.438 \pm 0.008$) while achieving a substantially lower Brier score ($0.0295$ vs.\ $0.0402$), reflecting its direct optimisation of probability calibration.
Hard BCE produces the worst calibration of the three ($\text{Brier} = 0.0455$), thus confirming that soft targets improve probability estimates regardless of the specific loss used.
Overall, soft supervision is strictly preferable to hard-label training: annotation fractions improve discrimination, and the MSE/Brier objective further improves calibration at no cost to PR-AUC.

\subsection{Effect of Demographic Conditioning}
\Cref{tab:task_b} reports Perspectivist Prediction macro PR-AUC for the text-only baseline and the three fusion strategies.
All three fusion strategies significantly outperform the text-only baseline: the additive model gains $+2.53$ pp ($+6.5\%$), early fusion $+2.34$ pp ($+6.0\%$), and concat-and-encode $+2.30$ pp ($+5.9\%$).
Demographic conditioning therefore provides a consistent and reliable improvement regardless of how deeply the demographic signal is integrated into the model.
Contrary to our initial expectations, deeper fusion does \emph{not} yield larger gains.
Instead, shallower integration depth is more beneficial (additive $\geq$ early fusion $\geq$ concat-and-encode), though pairwise differences are non-significant (early fusion vs.\ additive: $\Delta = -0.19$ pp, $[-0.7, +0.3]$; concat-and-encode vs.\ additive: $\Delta = -0.23$ pp, $[-0.8, +0.3]$).
We speculate this effect might be attributed to data sparsity: the per-group annotator counts in \textsc{P1SCO} are insufficient to exploit the additional capacity of deeper integration, while the zero-initialised additive residuals provide a strong inductive bias that prevents demographic corrections from overriding the text signal. See Appendix \ref{app:deltas} for detailed deltas per group.

\subsection{Prompting vs Fine-Tuning}
We compare our proposed architecture to model performance in zero- and few-shot settings using a mixture of open and closed-weight models of different sizes.
We also include abliterated versions of the models.
We evaluate all models on the \num{1104}-example test split of \textsc{P1SCO} under two ground truth conditions: a strict majority label (y) and a lenient any-annotator label ($s > 0$).
Prompts can be found in Appendix \ref{app:prompts}.
\Cref{tab:benchmark} shows detailed results per model and metric.
All models are well above the stratified random baseline ($\text{macro F1} = 0.10$) under the majority condition, with the exception of DeepSeek-R1-7B in zero-shot ($0.06$).
GPT-4o-mini zero-shot achieves the highest macro F1 under the majority condition ($0.36$), followed closely by Llama-3.1-8B-abliterated few-shot ($0.34$) and DeepSeek-R1-14B few-shot ($0.33$).
Few-shot prompting consistently outperforms zero-shot across models, and allowing the model to abstain with a \textit{none} label generally yields better results than forcing a prediction.
Abliterated models perform comparably to their non-abliterated counterparts, with Llama-3.1-8B-abliterated marginally outperforming the base Llama-3.1-8B in both conditions.
Under the any-annotator condition, rankings shift considerably: Mistral-7B-v0.3 emerges as the strongest model ($0.47$), suggesting it produces broader label sets that align well with minority annotator judgements.
Across conditions, our approach outperforms prompting methods. 

\begin{table*}[ht]
\centering
\small
\begin{tabular}{lcccccc}
\toprule
 & \multicolumn{3}{c}{Majority} & \multicolumn{3}{c}{Perspectivist} \\
\cmidrule(lr){2-4}\cmidrule(lr){5-7}
Model & Mac-F1 & Mic-F1 & PR-AUC & Mac-F1 & Mic-F1 & PR-AUC \\
\midrule
\multicolumn{7}{l}{\textit{Baselines}} \\
Random                     & 0.102          & 0.191          & 0.107          & 0.428          & 0.514          & 0.434          \\
Majority-class             & 0.000          & 0.000          & 0.103          & 0.308          & \textbf{0.607} & 0.428          \\
\citet{choi-etal-2020-ten} & \textbf{0.132} & \textbf{0.334} & \textbf{0.282} & \textbf{0.151} & 0.261          & \textbf{0.599} \\
\midrule
\multicolumn{7}{l}{\textit{Zero-shot}} \\
GPT-4o-mini       & \textbf{0.359} & \textbf{0.454} & \textbf{0.202} & 0.368          & 0.420          & 0.516          \\
DeepSeek-R1-32B   & 0.300          & 0.418          & 0.200          & 0.333          & 0.383          & 0.506          \\
DeepSeek-R1-14B   & 0.272          & 0.395          & 0.176          & 0.359          & 0.438          & 0.503          \\
DeepSeek-R1-7B    & 0.062          & 0.112          & 0.111          & 0.066          & 0.072          & 0.435          \\
Mistral-7B-v0.3   & 0.305          & 0.403          & 0.183          & \textbf{0.471} & \textbf{0.528} & \textbf{0.518} \\
Llama-3.1-8B      & 0.151          & 0.269          & 0.131          & 0.213          & 0.327          & 0.460          \\
Llama-3.1-8B-abl  & 0.296          & 0.417          & 0.194          & 0.364          & 0.414          & 0.508          \\
Qwen3-8B          & 0.259          & 0.327          & 0.133          & 0.331          & 0.389          & 0.454          \\
Qwen3-8B-abl      & 0.177          & 0.325          & 0.121          & 0.168          & 0.283          & 0.440          \\
\midrule
\multicolumn{7}{l}{\textit{Few-shot}} \\
GPT-4o-mini       & 0.306          & 0.428          & 0.199          & 0.271          & 0.316          & 0.497          \\
DeepSeek-R1-32B   & 0.329          & 0.446          & \textbf{0.211} & 0.405          & 0.460          & 0.523          \\
DeepSeek-R1-14B   & 0.334          & 0.451          & 0.196          & 0.409          & 0.492          & 0.512          \\
DeepSeek-R1-7B    & 0.225          & 0.331          & 0.146          & 0.265          & 0.322          & 0.464          \\
Mistral-7B-v0.3   & 0.332          & 0.447          & 0.206          & \textbf{0.460} & \textbf{0.515} & \textbf{0.529} \\
Llama-3.1-8B      & 0.291          & 0.446          & 0.176          & 0.284          & 0.398          & 0.481          \\
Llama-3.1-8B-abl  & \textbf{0.338} & \textbf{0.493} & 0.200          & 0.391          & 0.479          & 0.514          \\
Qwen3-8B          & 0.275          & 0.385          & 0.131          & 0.341          & 0.452          & 0.452          \\
Qwen3-8B-abl      & 0.249          & 0.372          & 0.153          & 0.287          & 0.426          & 0.461          \\
\midrule
\multicolumn{7}{l}{\textit{Fine-tuned (ours)}} \\
RoBERTa (Majority)                        & 0.420                    & \phantom{$^\dagger$}\textbf{0.564}$^\dagger$ & \phantom{$^\dagger$}\textbf{0.438}$^\dagger$ & 0.414          & 0.491          & \phantom{$^\dagger$}\textbf{0.711}$^\dagger$ \\
RoBERTa\textsubscript{text} (Perspectivist)    & 0.405                    & 0.531                    & 0.416                    & 0.410          & 0.497          & 0.389          \\
RoBERTa\textsubscript{+demo} (Perspectivist)   & \phantom{$^\dagger$}\textbf{0.425}$^\dagger$ & 0.552                    & 0.430                    & \textbf{0.422} & \textbf{0.505} & 0.414 \\
\bottomrule
\end{tabular}

\caption{Benchmark results on the test set (N$=$\num{1104}).
Majority uses the hard majority label; Perspectivist treats a label as positive if any annotator assigned it.
For each model, the best result across prompt conditions (none/no-none) is reported.
Abliterated models are marked -abl.
Bold: best within group; $\dagger$: best overall. See Appendix \ref{app:detailed_results} for more.}
\label{tab:benchmark}
\end{table*}

\section{Analysis}
\label{sec:analysis}
\subsection{Fusion Mechanism Insights}
\Cref{tab:task_b} and seed-level bootstrap confidence intervals confirm that all three fusion strategies significantly outperform the text-only perspectivist baseline, while pairwise differences between fusion strategies remain non-significant (early fusion vs.\ additive: $\Delta = -0.19$~pp, $[-0.7, +0.3]$; concat-then-encode vs.\ additive: $\Delta = -0.23$~pp, $[-0.8, +0.3]$).
The choice of fusion depth is therefore not the critical factor---what matters is that demographic information is included at all.

The additive model's competitive performance despite its architectural simplicity is informative.
Its zero-initialized weights allow the model to begin training as a text-only classifier and acquire demographic corrections only where annotation data supports them, acting as an implicit regularizer well-suited to the sparse per-group sample counts in P1SCO.
This feature might explain why deeper fusion does not translate into better generalisation.

At the label level, fusion gains cluster around the most semantically underspecified dimensions.
Power and Trust show the largest absolute improvements (\Cref{tab:per_label}), consistent with the intuition that these dimensions depend strongly on relational context and social positioning that demographic background helps calibrate.
Knowledge, Support, and Conflict, more directly recoverable from lexical content, see more modest gains.
Romance is the sole exception: gender alone outperforms the full gender--age--nationality triplet by +1.5pp (\Cref{tab:best_subset}), the only dimension for which adding age and nationality degrades performance.
We attribute this to the extreme sparsity of romance labels ($\approx$3\% prevalence), where nationality and age introduce statistical noise and sparsity that dilutes the gender signal.
Taken together, the per-label pattern shows that demographic conditioning is most valuable precisely where interpretation is most contested---thus validating the perspectivist framing of the task.

\subsection{Perspectivist Behaviour}
To examine whether demographic conditioning produces genuinely perspectival predictions rather than a uniform shift in label probabilities, we decompose macro PR-AUC gains by demographic subgroup.
For each annotator attribute (gender, age, nationality), we compare the text-only baseline against the additive model averaged across three seeds.
For gender, male annotators benefit considerably more from conditioning than female annotators (+4.6 pp vs.\ +0.7 pp).
Similarly for nationality, US annotators gain +4.1 pp against the text-only baseline, while UK annotators gain only +0.6 pp.
The most pronounced effect is for age: the 18-–24 group sees an average gain of +21.8 pp (from 42.4 to 64.1 macro PR-AUC), far exceeding any other age band, which all fall below +1.1 pp.
These patterns suggest that the text-only model already captures the perspectives of certain groups reasonably well, while demographic conditioning primarily helps groups whose readings diverge from that default.

For example, for the comment \textit{``That's funny because I've never seen the Colonel so closed minded''} the text-only model assigns a uniformly low probability of Power (p = $0.057$) and Trust (p = $0.058$) to all annotators, unable to differentiate between them.
The additive model, in turn, produces divergent predictions: a Male, 18–24, US annotator receives $p\_power = 0.985$ and $p\_trust = 0.953$ while Female annotators aged 35–44 from both the UK and the US, and a Male 45–54 UK annotator, receive $p\_power$ $\approx$ $0.15$ and $p\_trust$ $\approx$ $0.09$, consistent with their labels of 0.
The comment's implicit reference to authority and challenge appears to activate a power reading specifically for the younger US male, a perspectival response the model has learned to reproduce.

\paragraph{Comparison with LLMs.} To contextualise these findings, we compare our models against four strong LLMs (GPT-4o-mini, Mistral-7B, Llama-3.1-8B, DeepSeek-R1-14B; all few-shot) on per-group macro F1 agreement with each demographic group's majority labels.

Our additive fusion model achieves the highest raw F1 for every demographic group, confirming that explicit demographic conditioning is strictly more perspectivist than prompt-based elicitation.
This gap is widest for the 18--24 cohort, where fusion reaches $0.542$ vs.\ $0.368$ of the next-best model (Mistral-7B).
We find that LLMs exhibit their own implicit demographic alignments.
GPT-4o-mini is systematically most misaligned with 18--24 annotators ($\Delta = -0.048$) and with 60+ annotators ($\Delta = -0.026$); Mistral-7B shows the opposite tendency, over-representing 18--24 perspectives ($\Delta = +0.036$).
These skews are presumably inherited from pre-training and RLHF data distributions and cannot be corrected without retraining.
Our demographic-conditioned model, by contrast, makes its group-level assumptions explicit, auditable, and operationally distinct from uninstructed text processing.

In summary, the label- and model-level analyses suggest that the fusion model's differential gains represent \emph{intended perspectivism} where the model has learned to predict differently for groups that perceive social dimensions differently.

\section{Discussion}
\label{sec:discussion}

Our results confirm that demographic profiles carry genuine predictive signal: shuffle ablations show a $7.2$--$7.3$\% relative drop when demographic features are randomised (\Cref{tab:shuffle}), and the full gender–age–nationality triplet consistently outperforms any of its subsets (\Cref{tab:subset}).
However, this group-level signal is necessarily an approximation of individual interpretive behaviour.
Within any demographic group, substantial variation remains, a finding directly visible in the P1SCO annotation distributions.
\citet{orlikowski-etal-2023-ecological} warn against the ecological fallacy in annotation: inferring individual-level behaviour from group-level statistics.
Our framework is susceptible to exactly this fallacy when predictions are read as individual rather than population-level estimates.
We recommend that downstream applications treat model outputs as soft, distributional signals over demographic groups rather than as predictions about any particular annotator.

Demographic conditioning improves aggregate performance, but this comes with an inherent tension: a model that predicts differently for a 25--34-year-old woman from the UK than for a 55--64-year-old man from the US is making assumptions about individuals based on group membership and risks stereotyping.
We argue the distinction lies in intended use: demographic conditioning is appropriate when the goal is to describe the distribution of interpretations across a population, e.g., to understand how a piece of content will land across different communities.
However, it becomes problematic if it is used to make inferences about, or decisions affecting, specific individuals.
Our models should therefore be understood as tools for aggregate perspective modelling, not individual prediction.

While our findings support that demographic conditioning provides consistent and statistically significant improvements over text-only baselines, the absolute gains of approximately 2.5 percentage points appear small in isolation.
We note, however, that these gains are uniform across all three fusion strategies and survive shuffle ablations, indicating they reflect genuine demographic signal rather than noise.
Whether gains of this magnitude justify the additional complexity of demographic data collection and model conditioning in real-world deployments will depend on the application: for high-stakes settings where interpretive diversity matters, such as content moderation or public health communication, even small systematic improvements may be worthwhile, whereas for lower-stakes applications a text-only model may be sufficient.

\section{Related Work}
\label{sec:related}

\citet{choi-etal-2020-ten} introduced the ten-dimensional framework of social meaning we build on, demonstrating that dimensions such as knowledge, power, and support can be reliably detected from conversational text.
Subsequent work has applied this framework to study opinion dynamics on social media \citep{monti-etal-2022-language}, community well-being \citep{lucchini-etal-2022-reddit, aiello-etal-2021-epidemic}, and peer support in health contexts \citep{balsamo-etal-2023-pursuit}.
A key limitation of this line of work is its reliance on aggregated, majority-vote labels from a homogeneous annotator pool, obscuring the interpersonal variation.
A growing body of work challenges the assumption that annotation disagreement is noise to be resolved and argues that human label variation is a fundamental property of language data and calls for models that embrace rather than flatten it \cite[e.g.][]{plank-2022-problem,basile-etal-2021-need,uma-etal-2021-learning, ovesdotter-alm-2011-subjective}.
Motivated by this work, \citet{cercas-curry2026p1sco} provided disaggregated annotations of the ten social dimensions on social media comments.

Several studies have demonstrated that annotator background shapes subjective judgments in tasks such as hate speech detection, sentiment analysis, and politeness classification \citep{waseem-2016-racist, sap-etal-2022-annotators, goyal-etal-2022-your}.
In response, recent modelling work has explored incorporating annotator identity including gender, age, and political orientation as auxiliary signals, either through multi-task objectives or by conditioning classification heads on demographic embeddings \citep{davani-etal-2022-dealing, orlikowski-etal-2023-ecological,pei-jurgens-2023-annotator}.
Our fusion embedding framework extends this line of work by systematically comparing integration depth, from late additive adjustment to early concatenation before encoding, and by providing controlled shuffle ablations that isolate the genuine predictive contribution of demographic signals.

\section{Conclusion}
\label{sec:conclusion}

We introduced a perspectivist framework for social dimension prediction, establishing the first comprehensive benchmarks on P1SCO across zero-shot, few-shot, and fine-tuned paradigms.
We show that annotator demographic profiles carry genuine, non-spurious predictive signal: demographic-conditioned fusion embeddings yield consistent and statistically significant improvements over text-only baselines across all three fusion strategies (+$5.9$--$6.5$\% relative macro PR-AUC), confirmed by shuffle ablations that return performance to the text-only range when demographic assignments are broken.
Contrary to our expectations, shallower fusion is as effective as deeper integration, a result we attribute to the sparsity of per-group annotations in P1SCO and the strong inductive bias of additive zero-initialised residuals.
Per-label analysis reveals that gains concentrate on the most semantically underspecified dimensions---Power and Trust---where shared demographic experience most strongly mediates interpretation, while Romance uniquely benefits from gender alone.
Subgroup analysis further shows that conditioning disproportionately benefits groups whose perspectives diverge from the text-only default.

Future work should pursue richer and intersectional demographic representations, cross-dataset validation, and methods that move beyond group-level stereotyping towards more individual-aware perspective modelling.

\section*{Limitations}
While our results demonstrate the value of demographic conditioning for perspectivist social dimension prediction, several limitations bound the scope of our conclusions:

\paragraph{Annotation sparseness}
Per-group annotator counts in P1SCO are uneven, with some demographic combinations represented by very few participants.
We attribute the failure of deeper fusion strategies to outperform additive fusion partly to this sparsity: the additional model capacity offered by early and concat-then-encode fusion cannot be fully exploited when group-level signal is thin.
Richer modelling of intersectional identities in particular will require datasets with substantially larger per-group annotator counts.

\paragraph{Limited demographic variables studied}
We restricted ourselves to three demographic variables (gender, age, nationality).
Personality traits, political orientation, and education exist in the dataset but are not currently modelled.
We are also limit our study to binary gender.
Moreover, the dataset only includes data in English and participants from the UK and US so our results may not generalise across languages and cultures.
Extending perspectivist social dimension modelling to multilingual and cross-cultural settings remains an important direction for future work.

\paragraph{Evaluation setup}
We test our proposed architecture on only one dataset.
There's no cross-dataset validation, so it is unknown whether the demographic conditioning gains generalise to other social dimension datasets or annotation setups.

\paragraph{Practical significance}
PR-AUC gains are modest in absolute terms, although results are consistent across fusion strategies and robust to shuffle ablations.

\section*{Acknowledgments}
AI-based assistants were used to support code development and provide writing and editing assistance during manuscript preparation.
All content, analyses, and conclusions were reviewed and verified by the authors.

\bibliography{custom}

\appendix

\section{Prompts}
\label{app:prompts}

\begin{tcolorbox}[
  breakable,
  colback=gray!10,
  colframe=gray!50,
  title=Prompt template (zero-shot),
  fonttitle=\bfseries\small,
  fontlower=\small,
  left=6pt, right=6pt, top=6pt, bottom=6pt
]
\small
You are an expert text annotation tool.
You output only valid JSON.
Never explain, never add commentary.

\medskip
Annotate the text below for the following social dimensions.

\medskip
\textbf{Dimensions:}

\smallskip
\begin{tabular}{@{}lp{0.68\linewidth}@{}}
\textbf{knowledge}  & \raggedright Exchange of ideas or information; learning, teaching, asking or answering questions \tabularnewline[3pt]
\textbf{power}      & \raggedright Having power over another's behaviour; commands, coercion, dominance, rule enforcement \tabularnewline[3pt]
\textbf{status}     & \raggedright Conferring appreciation, admiration, gratitude, or prestige upon another \tabularnewline[3pt]
\textbf{trust}      & \raggedright Willingness to rely on another's actions or judgements; expressing confidence or faith \tabularnewline[3pt]
\textbf{support}    & \raggedright Giving emotional or practical aid, encouragement, care, or companionship \tabularnewline[3pt]
\textbf{similarity} & \raggedright Shared interests, motivations, outlooks, or alignment of perspectives \tabularnewline[3pt]
\textbf{identity}   & \raggedright Shared sense of belonging to the same community, culture, or group \tabularnewline[3pt]
\textbf{fun}        & \raggedright Leisure, laughter, humour, playful teasing, entertainment, or joy \tabularnewline[3pt]
\textbf{conflict}   & \raggedright Disagreement, criticism, frustration, opposition, or hostility \tabularnewline[3pt]
\textbf{romance}    & \raggedright Intimacy or expressions related to a sentimental or sexual relationship \tabularnewline[3pt]
\textbf{none}       & \raggedright No other dimension is clearly present \tabularnewline
\end{tabular}

\medskip
\textbf{Rules:}
\begin{itemize}[noitemsep, topsep=2pt, leftmargin=*]
  \item Only label a dimension if clearly expressed.
  \item Multiple labels are allowed.
  \item Never combine \texttt{none} with other labels.
  \item Output only JSON. No explanation. No markdown.
\end{itemize}

\medskip
\textbf{Output format:} 

\texttt{\{\{"labels": ["label1", "label2"]\}\}}

\textbf{Text:} \texttt{\{candidate\_text\}}
\normalsize
\end{tcolorbox}

\begin{tcolorbox}[
  breakable,
  colback=gray!10,
  colframe=gray!50,
  title=Prompt template (few-shot),
  fonttitle=\bfseries\small,
  left=8pt, right=8pt, top=6pt, bottom=6pt
]
\small
You are an expert annotator of social dimensions.
Label the text with ALL applicable dimensions.

\medskip
\textbf{Dimensions:}

\smallskip
\begin{tabular}{@{}lp{0.68\linewidth}@{}}
\textbf{knowledge}  & \raggedright Exchange of ideas or information; learning, teaching, asking or answering questions \tabularnewline[3pt]
\textbf{power}      & \raggedright Having power over another's behaviour; commands, coercion, dominance, rule enforcement \tabularnewline[3pt]
\textbf{status}     & \raggedright Conferring appreciation, admiration, gratitude, or prestige upon another \tabularnewline[3pt]
\textbf{trust}      & \raggedright Willingness to rely on another's actions or judgements; expressing confidence or faith \tabularnewline[3pt]
\textbf{support}    & \raggedright Giving emotional or practical aid, encouragement, care, or companionship \tabularnewline[3pt]
\textbf{similarity} & \raggedright Shared interests, motivations, outlooks, or alignment of perspectives \tabularnewline[3pt]
\textbf{identity}   & \raggedright Shared sense of belonging to the same community, culture, or group \tabularnewline[3pt]
\textbf{fun}        & \raggedright Leisure, laughter, humour, playful teasing, entertainment, or joy \tabularnewline[3pt]
\textbf{conflict}   & \raggedright Disagreement, criticism, frustration, opposition, or hostility \tabularnewline[3pt]
\textbf{romance}    & \raggedright Intimacy or expressions related to a sentimental or sexual relationship \tabularnewline[3pt]
\textbf{none}       & \raggedright No dimension is clearly present \tabularnewline
\end{tabular}

\medskip
\textbf{Rules:}
\begin{itemize}[noitemsep, topsep=2pt, leftmargin=*]
  \item Only label a dimension if clearly expressed.
  \item Multiple labels are allowed.
  \item Never combine \texttt{none} with other labels.
  \item If no dimension is present, return an empty list.
  \item Output ONLY valid JSON. No explanation.
\end{itemize}

\medskip
\textbf{Output format:} 

\texttt{\{"labels": ["label1", "label2"]\}}

\medskip
\textbf{Examples:}

\smallskip
\begin{tabular}{@{}p{0.62\linewidth}p{0.32\linewidth}@{}}
\textit{``I'd recommend using canned beer instead of bottled beer the first few times.''} & \texttt{["knowledge"]} \tabularnewline[4pt]
\textit{``You must submit your report by noon or there will be consequences.''} & \texttt{["power"]} \tabularnewline[4pt]
\textit{``You did an amazing job on this project!''} & \texttt{["status"]} \tabularnewline[4pt]
\textit{``I trust your judgment—go ahead with the plan.''} & \texttt{["trust"]} \tabularnewline[4pt]
\textit{``I'm really sorry you're going through this. I'm here for you.''} & \texttt{["support"]} \tabularnewline[4pt]
\textit{``I totally agree with your views on this topic.''} & \texttt{["similarity"]} \tabularnewline[4pt]
\textit{``As fellow artists, we understand the struggle.''} & \texttt{["identity"]} \tabularnewline[4pt]
\textit{``Haha that's hilarious, you made my day!''} & \texttt{["fun"]} \tabularnewline[4pt]
\textit{``Your argument makes no sense and you're ignoring the facts.''} & \texttt{["conflict"]} \tabularnewline[4pt]
\textit{``I think about you all the time, you make my days brighter.''} & \texttt{["romance"]} \tabularnewline[4pt]
\textit{``I totally agree with you, and you explained it really well.''} & \texttt{["similarity", "status"]} \tabularnewline[4pt]
\textit{``The sky is blue.''} & \texttt{["none"]} \tabularnewline
\end{tabular}

\medskip
\hrule
\medskip

\textbf{Now classify:}

\medskip
\textbf{Text:} \texttt{\{candidate\_text\}}
\normalsize
\end{tcolorbox}

\newpage
\onecolumn
\section{Detailed results per dimension}
\label{app:detailed_results}
\begin{table*}[h]
\centering
\small
\setlength{\tabcolsep}{4pt}
\begin{tabular}{lcccccccccc}
\toprule
Model & Know. & Power & Status & Trust & Supp. & Sim. & Ident. & Fun & Conf. & Rom. \\
\midrule
\multicolumn{11}{l}{\textit{Baselines}} \\
Random & 0.330 & 0.009 & 0.098 & 0.013 & 0.226 & 0.154 & 0.059 & 0.086 & 0.094 & 0.009 \\
Majority-class & 0.331 & 0.008 & 0.089 & 0.014 & 0.206 & 0.155 & 0.062 & 0.059 & 0.102 & 0.010 \\
\midrule
\multicolumn{11}{l}{\textit{Zero-shot}} \\
GPT-4o-mini & 0.486 & 0.019 & 0.254 & 0.014 & 0.376 & 0.228 & 0.122 & 0.130 & 0.350 & 0.044 \\
DeepSeek-R1-32B & 0.490 & 0.025 & 0.203 & 0.014 & 0.365 & 0.186 & 0.097 & 0.113 & 0.373 & 0.138 \\
DeepSeek-R1-14B & 0.462 & 0.019 & 0.135 & 0.013 & 0.302 & 0.202 & 0.080 & 0.149 & 0.332 & 0.068 \\
DeepSeek-R1-7B & 0.364 & 0.008 & 0.089 & 0.014 & 0.217 & 0.158 & 0.062 & 0.069 & 0.108 & 0.017 \\
Mistral-7B-v0.3 & 0.477 & 0.027 & 0.184 & 0.036 & 0.351 & 0.204 & 0.081 & 0.114 & 0.289 & 0.067 \\
Llama-3.1-8B & 0.400 & 0.049 & 0.097 & 0.014 & 0.276 & 0.167 & 0.061 & 0.094 & 0.140 & 0.010 \\
Llama-3.1-8B-abl & 0.478 & 0.024 & 0.133 & 0.020 & 0.391 & 0.180 & 0.094 & 0.132 & 0.362 & 0.121 \\
Qwen3-8B & 0.403 & 0.008 & 0.149 & 0.014 & 0.266 & 0.185 & 0.070 & 0.072 & 0.149 & 0.010 \\
Qwen3-8B-abl & 0.382 & 0.012 & 0.088 & 0.014 & 0.208 & 0.163 & 0.066 & 0.115 & 0.153 & 0.010 \\
\midrule
\multicolumn{11}{l}{\textit{Few-shot}} \\
GPT-4o-mini & 0.459 & 0.008 & 0.246 & 0.014 & 0.374 & 0.188 & 0.124 & 0.131 & 0.380 & 0.062 \\
DeepSeek-R1-32B & 0.535 & 0.022 & 0.211 & 0.014 & 0.373 & 0.248 & 0.124 & 0.149 & 0.365 & 0.067 \\
DeepSeek-R1-14B & 0.473 & 0.060 & 0.184 & 0.018 & 0.345 & 0.223 & 0.110 & 0.154 & 0.308 & 0.084 \\
DeepSeek-R1-7B & 0.454 & 0.010 & 0.105 & 0.014 & 0.263 & 0.191 & 0.080 & 0.103 & 0.137 & 0.096 \\
Mistral-7B-v0.3 & 0.523 & 0.017 & 0.179 & 0.018 & 0.403 & 0.238 & 0.095 & 0.124 & 0.346 & 0.119 \\
Llama-3.1-8B & 0.468 & 0.047 & 0.142 & 0.019 & 0.401 & 0.182 & 0.094 & 0.119 & 0.258 & 0.032 \\
Llama-3.1-8B-abl & 0.458 & 0.027 & 0.203 & 0.016 & 0.394 & 0.245 & 0.088 & 0.138 & 0.376 & 0.054 \\
Qwen3-8B & 0.410 & 0.020 & 0.138 & 0.014 & 0.224 & 0.176 & 0.072 & 0.106 & 0.136 & 0.010 \\
Qwen3-8B-abl & 0.436 & 0.022 & 0.109 & 0.014 & 0.295 & 0.170 & 0.075 & 0.165 & 0.238 & 0.010 \\
\midrule
\multicolumn{11}{l}{\textit{Fine-tuned (ours)}} \\
RoBERTa (Majority) & \textbf{0.773} & \textbf{0.170} & \textbf{0.446} & 0.064 & \textbf{0.715} & \textbf{0.435} & 0.195 & 0.512 & 0.702 & \textbf{0.372} \\
RoBERTa\textsubscript{text} (Perspectivist) & 0.762 & 0.083 & 0.425 & 0.061 & 0.699 & 0.414 & \textbf{0.206} & 0.493 & 0.671 & 0.346 \\
RoBERTa\textsubscript{+demo} (Perspectivist) & \textbf{0.773} & 0.160 & 0.422 & \textbf{0.066} & 0.702 & 0.431 & 0.197 & \textbf{0.520} & \textbf{0.706} & 0.325 \\
\bottomrule
\end{tabular}
\caption{Per-label PR-AUC on the test set (Majority task) for all models. Bold: best per column. }
\label{tab:per_label_maj}
\end{table*}
 
\begin{table*}[h]
\centering
\small
\setlength{\tabcolsep}{4pt}
\begin{tabular}{lcccccccccc}
\toprule
Model & Know. & Power & Status & Trust & Supp. & Sim. & Ident. & Fun & Conf. & Rom. \\
\midrule
\multicolumn{11}{l}{\textit{Baselines}} \\
Random & 0.733 & 0.216 & 0.469 & 0.317 & 0.621 & 0.646 & 0.528 & 0.325 & 0.346 & 0.116 \\
Majority-class & 0.730 & 0.217 & 0.471 & 0.312 & 0.613 & 0.652 & 0.517 & 0.327 & 0.332 & 0.102 \\
\midrule
\multicolumn{11}{l}{\textit{Zero-shot}} \\
GPT-4o-mini & 0.813 & 0.267 & 0.614 & 0.322 & 0.727 & 0.690 & 0.530 & 0.502 & 0.539 & 0.159 \\
DeepSeek-R1-32B & 0.801 & 0.270 & 0.573 & 0.316 & 0.743 & 0.658 & 0.536 & 0.474 & 0.490 & 0.199 \\
DeepSeek-R1-14B & 0.824 & 0.267 & 0.515 & 0.366 & 0.728 & 0.664 & 0.536 & 0.500 & 0.426 & 0.199 \\
DeepSeek-R1-7B & 0.748 & 0.224 & 0.473 & 0.312 & 0.617 & 0.655 & 0.519 & 0.350 & 0.345 & 0.112 \\
Mistral-7B-v0.3 & 0.821 & 0.318 & 0.554 & 0.352 & 0.754 & 0.672 & 0.552 & 0.438 & 0.559 & 0.162 \\
Llama-3.1-8B & 0.762 & 0.254 & 0.474 & 0.312 & 0.720 & 0.653 & 0.531 & 0.426 & 0.363 & 0.110 \\
Llama-3.1-8B-abl & 0.824 & 0.316 & 0.511 & 0.343 & 0.727 & 0.671 & 0.526 & 0.475 & 0.533 & 0.152 \\
Qwen3-8B & 0.756 & 0.232 & 0.487 & 0.313 & 0.660 & 0.689 & 0.537 & 0.370 & 0.360 & 0.132 \\
Qwen3-8B-abl & 0.761 & 0.232 & 0.476 & 0.315 & 0.637 & 0.655 & 0.518 & 0.341 & 0.360 & 0.106 \\
\midrule
\multicolumn{11}{l}{\textit{Few-shot}} \\
GPT-4o-mini & 0.798 & 0.236 & 0.596 & 0.317 & 0.688 & 0.681 & 0.527 & 0.450 & 0.524 & 0.152 \\
DeepSeek-R1-32B & 0.806 & 0.248 & 0.578 & 0.312 & 0.732 & 0.714 & 0.557 & 0.520 & 0.550 & 0.208 \\
DeepSeek-R1-14B & 0.828 & 0.260 & 0.575 & 0.335 & 0.745 & 0.694 & 0.545 & 0.489 & 0.450 & 0.199 \\
DeepSeek-R1-7B & 0.791 & 0.233 & 0.480 & 0.314 & 0.665 & 0.683 & 0.540 & 0.401 & 0.375 & 0.158 \\
Mistral-7B-v0.3 & 0.832 & 0.283 & 0.570 & 0.336 & 0.753 & 0.711 & 0.567 & 0.469 & 0.599 & 0.174 \\
Llama-3.1-8B & 0.835 & 0.240 & 0.507 & 0.315 & 0.728 & 0.658 & 0.541 & 0.456 & 0.412 & 0.118 \\
Llama-3.1-8B-abl & 0.834 & 0.293 & 0.548 & 0.315 & 0.714 & 0.723 & 0.529 & 0.487 & 0.510 & 0.191 \\
Qwen3-8B & 0.765 & 0.229 & 0.512 & 0.312 & 0.646 & 0.674 & 0.531 & 0.386 & 0.348 & 0.115 \\
Qwen3-8B-abl & 0.790 & 0.241 & 0.508 & 0.314 & 0.698 & 0.678 & 0.533 & 0.378 & 0.374 & 0.102 \\
\midrule
\multicolumn{11}{l}{\textit{Fine-tuned (ours)}} \\
RoBERTa (Majority) & \textbf{0.943} & \textbf{0.440} & \textbf{0.794} & \textbf{0.523} & \textbf{0.905} & \textbf{0.855} & \textbf{0.692} & \textbf{0.720} & \textbf{0.818} & \textbf{0.416} \\
RoBERTa\textsubscript{text} (Perspectivist) & 0.690 & 0.132 & 0.412 & 0.157 & 0.607 & 0.434 & 0.269 & 0.404 & 0.561 & 0.226 \\
RoBERTa\textsubscript{+demo} (Perspectivist) & 0.701 & 0.200 & 0.436 & 0.205 & 0.620 & 0.456 & 0.296 & 0.440 & 0.583 & 0.209 \\
\bottomrule
\end{tabular}
\caption{Per-label PR-AUC on the test set (Perspectivist condition) for all models. Bold: best per column. For LLMs and Majority task, Perspectivist uses candidate-level any-annotator labels ($s > 0$); for Perspectivist task, annotation-level evaluation is used (individual annotator labels).}
\label{tab:per_label_persp}
\end{table*}

\newpage
\onecolumn
\section{Detailed results per socio-demographic subgroup}
\label{app:deltas}
\begin{figure*}[ht]
    \centering
    \includegraphics[width=1\linewidth]{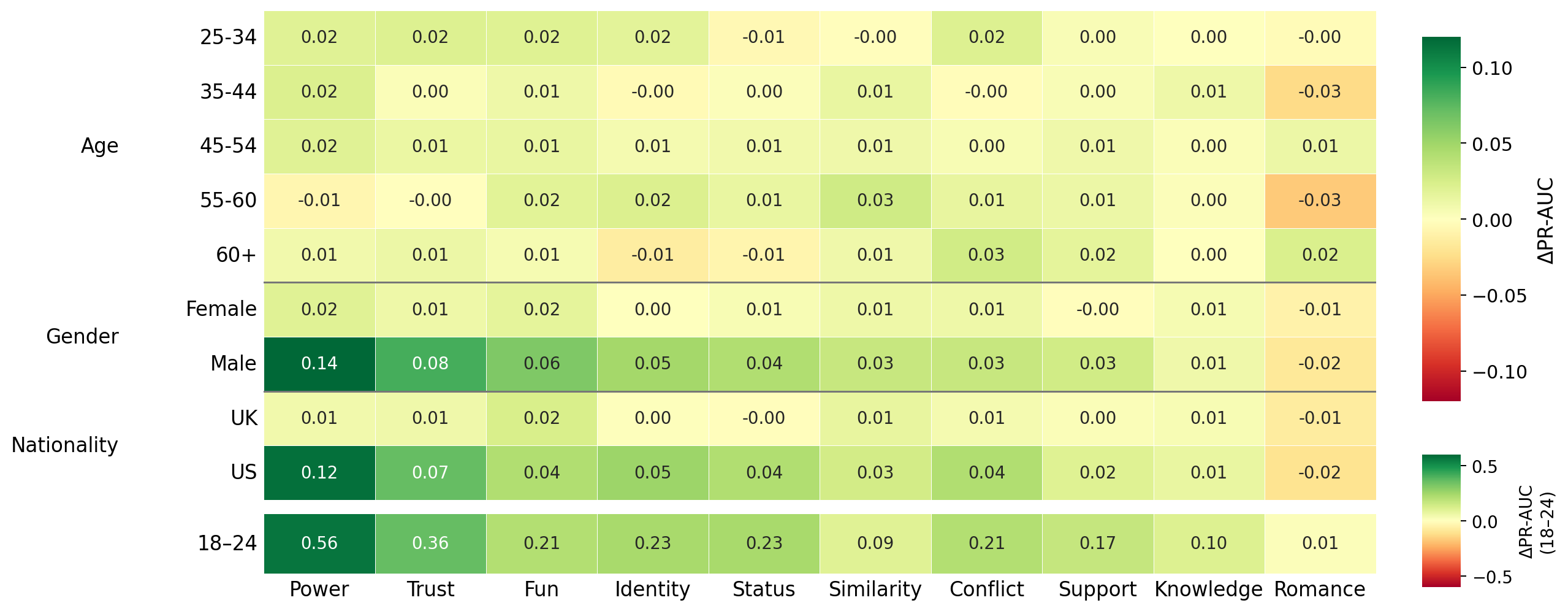}
    \caption{Per-label PR-AUC delta (additive vs. text-only baseline) broken down by demographic subgroup. }
    \label{fig:delta_heatmap}
\end{figure*}

\end{document}